\documentclass{easychair}
\usepackage{cite}
\usepackage{url}
\usepackage{hyperref}
\usepackage{doi}
\usepackage{amsmath,amssymb,amsfonts}
\usepackage{graphicx}
\usepackage{lipsum}  
\usepackage{textcomp}
\usepackage{xcolor}
\usepackage{caption}
\usepackage{subcaption}
\usepackage{tikz} 
\usepackage{algorithm}
\usepackage{algpseudocode}

\usepackage{float}
\usetikzlibrary{arrows.meta, positioning}

\tikzset{
  box/.style={rectangle, draw, rounded corners, minimum width=0cm, minimum height=1cm, align=center, font=\small},
  arrow/.style={-Latex, thick},
}
\usepackage{doc}

\title{Novel Architecture of RPA In Oral Cancer Lesion Detection}

\author{
Revana Magdy  \href{https://orcid.org/0009-0004-7644-537X}{0009-0004-7644-537X}, Joy Naoum \href{https://orcid.org/0009-0005-8314-2702}{0009-0005-8314-2702} , Ali Hamdi \href{https://orcid.org/0000-0002-2301-6588}{0000-0002-2301-6588},
 $^{1}$  
\texttt{\{revana.magdy} ,
$^{2}$
\texttt{joy.ehab} 
 ,
$^{3}$ 
\texttt{alihamdif\}@msa.edu.eg}
}
\titlerunning{Novel Architecture of RPA in oral cancer lesion detection}
\institute{MSA University, Giza, Egypt}
\begin{document}
\maketitle
\vspace{2\baselineskip} 

\begin{center}
\textbf{Abstract}
\end{center}

\begin{quote}
\small 
\setlength{\leftskip}{1cm}
\setlength{\rightskip}{1cm}
Accurate and early detection of oral cancer lesions is crucial for effective diagnosis and treatment. This study evaluates two RPA implementations, OC-RPAv1 and OC-RPAv2, using a test set of 31 images. OC-RPAv1 processes one image per prediction in an average of 0.29 seconds, while OC-RPAv2 employs a Singleton design pattern and batch processing, reducing prediction time to just 0.06 seconds per image. This represents a 60-100x efficiency improvement over standard RPA methods, showcasing that design patterns and batch processing can enhance scalability and reduce costs in oral cancer detection. \\

\end{quote}


\section{Introduction}
\label{sect:introduction}Improving patient survival rates and outcomes begins with the early and precise detection of oral cancer. Although diagnostic imaging technologies continue to advance, the clinical workflows for the analysis of oral lesions remain challenged by subjective human interpretation, delays in workflow, and inconsistent human decisions. Such factors can result in diagnostic delays and can hamper clinical decision-making efficiency. Thus, the demand for automated intelligent systems that can accelerate the diagnostic workflow without compromise to determinism and accuracy is justified.  

Recent automation innovations in the form of Robotic Process Automation (RPA) have emerged as a novel automation technique in the optimisation of healthcare workflows. RPA contributes to more efficient and reliable workflow automation by taking over tedious and rule-bound processes that would otherwise require human effort. Healthcare professionals within RPA-supported frameworks become workflow automation designers, performing the diagnostic sequence by passing diagnostic variables through an automated system. The healthcare professional will perform the diagnosis in a computationally optimal sequence and will feed the variables into the automated system, thus bridging the workflow automation gap between the automated system and the clinical professional.

RPA has been employed across several industries for automation, data extraction, and workflow management. In healthcare, it has been adopted mainly to automate image processing, laboratory data management, and patient data analysis. Automating these tasks alleviates operational burdens, drastically reducing human error. This not only enhances clinical practice consistency, but efficiency as well. On the other hand, automation through programming affords greater flexibility for deep learning model integration, data pipeline optimisation, and third-party libraries for model deployment. As much as RPA systems aim for simplicity, automation through CNN models provides greater control, extensive scalability, and sophisticated optimisation. The combination of the RPA and CNN model automation methods presents a usable and efficient framework as each complements the other.

This study develops an automated system for detecting oral cancer lesions using a deep learning model. It employs a CNN architecture, EfficientNetV2B1, trained on a clinical dataset with sixteen types of oral cancer lesions, achieving strong predictive results.

RPA is utilized across various industries for automation, particularly in healthcare where it automates tasks like image processing and patient data analysis. This reduces human error and improves clinical efficiency. In contrast, programming-driven automation allows for deep learning model integration and better data optimization. Combining RPA and CNN model automation creates an efficient framework, leveraging the strengths of both methods.

This study implements the other frameworks of automation around a deep learning model to create an automated system for oral cancer lesion detection. This system utilizes a CNN architecture, EfficientNetV2B1, which has been trained on a clinical dataset The dataset is organized into four major diagnostic classes(Healthy, Benign, OPMD, and Oral Cancer)ach of which includes multiple types. This model has yielded strong predictive outcomes.

This study marks the first work on:
\begin{itemize}
    \item  Integrating Singleton and Batch Processing design patterns into an optimized process to substantially reduce computation time. 
    \item  Achieving a 60–100× improvement in performance relative to RPA-based deployments while preserving diagnostic accuracy.
    \item  Achieving predictions 60-90× faster than conventional RPA systems while preserving predictions.
    \item  Developing economically and technologically scalable systems in actual healthcare to enhance resource utilization for faster detection and more efficient systems.
 .  
\end{itemize}

\subsection{RPA Workflow}
\begin{figure}[htbp]

\centering
\begin{tikzpicture}
\node[anchor=south west, inner sep=1] (img) at (0,0) {\includegraphics[width=0.5\textwidth]{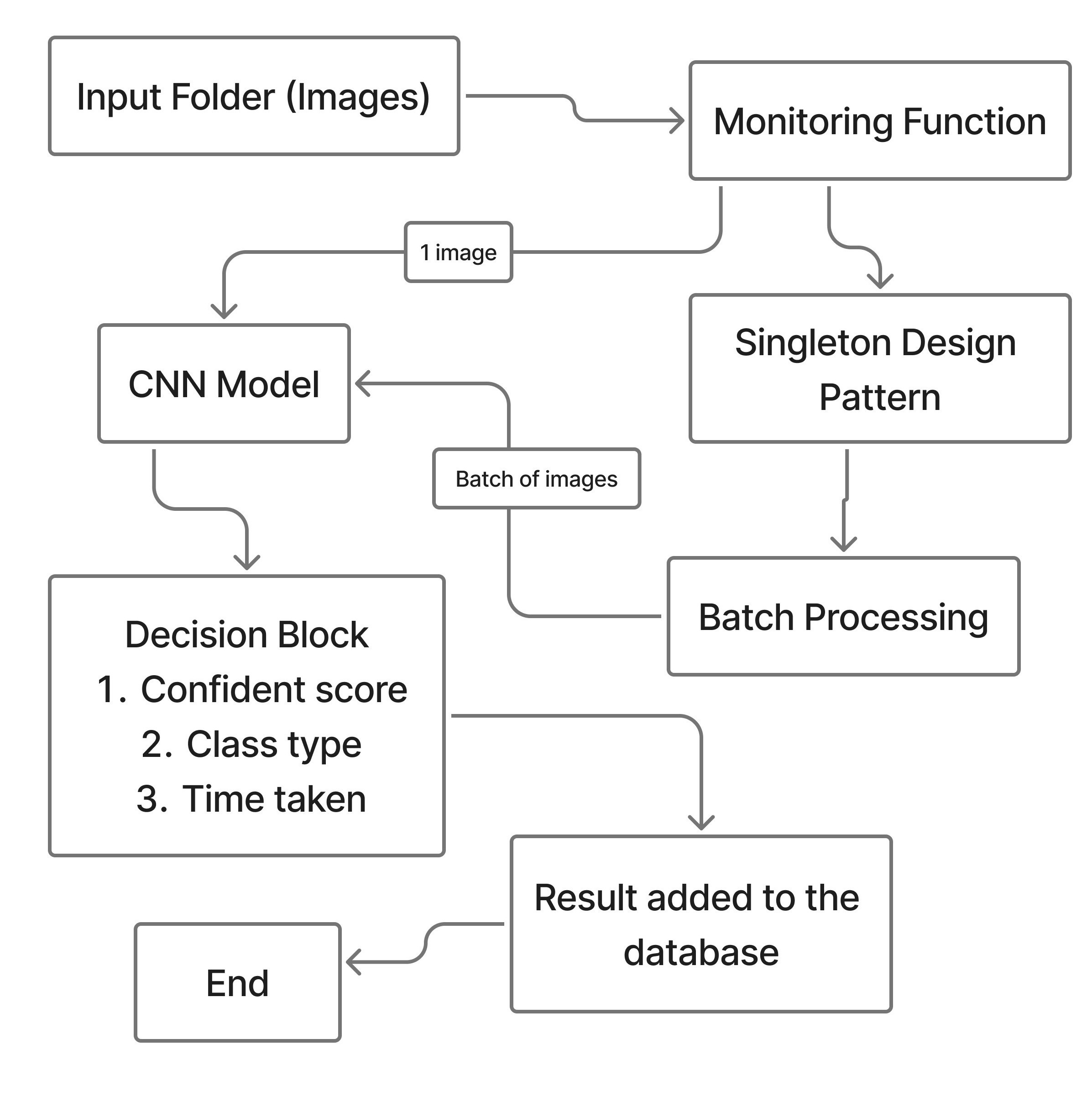}};
\end{tikzpicture}
\caption{The OC RPA workflow illustrates two parallel pipelines that converge at the same CNN model. The first pipeline processes a single image through the CNN model, while the second applies design patterns—specifically, the Singleton and Batch Strategy patterns—to process a batch of images.}
\label{fig:Workflow}
\end{figure}
This study evaluates robotic process automation by combining Singleton and Batch Processing design patterns within the CNN prediction workflow to improve cost and time efficiency. The system utilizes UiPath to manage the automation pipeline, invoking a Python function that loads the EfficientNetV2B1 model once and keeps it in memory. Each image is processed through a prediction function that logs and stores results automatically, with processed files moved to a separate directory to ensure data integrity. This workflow facilitates effective communication between RPA logic and AI inference, enabling scalable and automated lesion classification.

~\ref{sect:Related Work} Related Work in Implementing RPA in Healthcare, while ~\ref{sec:methodology} discusses the dataset, preprocessing, augmentation, and the proposed model architecture. Section ~\ref{sec:experiments} details the experimental setup and results, which are discussed in Section ~\ref{sec:discussion}. Section ~\ref{sec:conclusion} concludes the study and highlights core findings, and Section ~\ref{sec:future work} presents possible avenues for future work. OC RPA workflow is shown in Figure ~\ref{fig:Workflow}.

\section{Related Work}
\label{sect:Related Work}
RPA is used to automate some of the mundane, rule-driven activities within the healthcare domain. For example, information systems in hospitals utilize RPA to minimize operational manual work and human error as highlighted in the research conducted by Park et al. \cite{ref16}. However, the study also pointed out issues concerning exception management, interoperability of software, and compliance with laws and regulations.  

Zhou et al. \cite{ref20} designed a framework on RPA technology for the automation of the processes of preparing and managing clinical images for diagnostics. Although automation systems can reduce the amount of work that needs to be done manually, performing automation on large volumes of diagnostic images, especially high-resolution ones, leads to inefficiencies, and challenges of scale and real time operations arise. The study by Venigandla and Tatikonda \cite{ref18} on the integration of RPA with deep learning highlighted that UiPath, a low code RPA automation integration, is user friendly but computational heavy tasks are inefficient in terms of resource use. The study also pointed out that there are considerable resource gains to be made when Python is used for batch image processing in conjunction with automated model inference.

Abdellaif et al. \cite{ref9} presented the LMV-RPA technique and showed how augmented automation with Python can complement standard RPA in improving clinical workflows by reducing clinical errors and improving performance. The study showed the usefulness of the design pattern in Batch Processing for simultaneous multiple input handling and the Singleton pattern in model loading. This hybrid RPA-Python approach lessens sequential execution bottlenecks in low-code environments. 

Kim et al. \cite{ref11} helped extend the work by using RPA and Python for the automation of the cancer detection pathology images workflow and reported higher speed in workflow execution in comparison to RPA-only workflows. This was due to the flexible nature of Python for implementing numerous preprocessing, augmentation, and batch inference techniques. Li and Wang \cite{ref12} hinged computing intensive tasks on Python and RPA drove workflow orchestration in their hybrid structure. The work yielded improved repeatability, error propagation, and enhanced throughput on large scale medical datasets.

The adoption of RPA and deep learning technologies in healthcare is on the rise.

Multiple recent reports indicate that the automation of complex workflows, such as the multi-class lesion identification and diagnostic report generation pipelines, is achievable using RPA for workflow orchestration and Python for interpreting algorithmic inferences. These hybrid automation systems utilize the best features of the two approaches. RPA handles workflow standardization, error tracking, and ease of deployment, while Python scripts facilitate optimized computation, parallel processing, and advanced model management. However, in high-throughput clinical settings, the challenges of system complexity, execution speed, and cost remain unresolved. Swetha et al. \cite{ref19} discussed the RPA applications in healthcare and its potential for improving efficacy and diminishing administrative workload in detail. The study underscored the need for RPA to be integrated into the healthcare IT systems already in use. Things like automated patient scheduling, billing, and claim management are practical applications. Automation enhanced the operational and user experience of FloridaBlue.com as detailed in Esther \cite{ref7}, which integrated Python and Sikuli for the user-end automation, while the operational side All these studies show a unified trend: RPA platforms ease workflow management while decreasing human job roles. Nevertheless, the best automation in medical imaging, which is fast, profitable, and cost-efficient, also includes the use of Python and strong design principles. This incorporated approach streamlines rapid clinical decision-making and diagnostic turnaround times, allowing for more expansive healthcare systems. In their extensive literature review, Guarda et al. \cite{ref17} also analyzed RPA use in healthcare specifically within the domains of radiology, appointment systems, and data management. They found that RPA not only enhanced productivity and precision but also detected chronic problems related to unsophisticated EMR systems.  

The results from these studies are quite similar to our benchmark study on RPA-inspired pipelines for detecting oral cancer, particularly with respect to the balance between the rate of automation and clinical acceptability. In his review on the use of AI for the diagnosis of oral cancer and oral potentially malignant disorders clinical images, Mirfendereski et al. \cite{ref14} raised important points regarding multiclass classification and the consolidation of the diagnostic processes, both of which are automation concerns addressed in our study. Baweja \cite{ref5,ref6} compared the three top RPA platforms—UiPath, Automation Anywhere, and Blue Prism—assessing their architecture, automation features and domain appropriateness for each. It examined the significance of intelligent automation particularly in healthcare and media and provided guidance on tool selection as a function of automation scalability, integration flexibility and complexity of the tasks. The documents cited provide the basis for our benchmarking study of RPA-style automation pipelines in the detection of oral cancer lesions, where the efficiency of execution and the clinical relevance are of utmost importance.

In the proposed research work, the focus was on benchmarking the CLASEG framework provided by Al-Ali et al. \cite{ref3}, for integrated multi-class classification and segmentation. CLASEG deep learning framework integrated multi-class classification and segmentation with the differential diagnosis of oral lesions. It identified 16 various lesion types, crossing different regions of the mouth, and combined pixel segmentation with classification at the diagnostic level to improve clinical precision. It served the redesigned EfficientNetV2B1-based pipeline that we use for the current research.

\subsection{Relationship Between RPA and AI in the Proposed Workflow}
Nividous \cite{ref15} differentiates \textbf{Robotic Process Automation (RPA)} and Artificial Intelligence (AI) considering their respective and complementary functionalities within an intelligent automation framework. RPA is limited to automation of mundane, rote, and structured digital processes like data entry and file manipulation, and does this without any understanding of the data. On the other hand, AI renders a system the ability to learn and mimic some of the higher human cognitive functions, as it can reason and predict, with ‘data’ which can be structured and unstructured.

\begin{table*}[!ht]
\centering
\caption{Mapping of RPA and AI Concepts to the Proposed Oral Cancer Detection Workflow}
\resizebox{\textwidth}{!}{
\begin{tabular}{|p{3.2cm}|p{5.3cm}|p{7.5cm}|}
\hline
\textbf{Concept} & \textbf{General Definition (Nividous, 2025)} & \textbf{Application in This Project} \\ \hline
\textbf{RPA Purpose} & Automates structured, rule-based processes without cognitive understanding. & UiPath Studio and Automation Anywhere automate the inference workflow—loading medical images, executing the trained CNN model, and logging prediction time. \\ \hline
\textbf{AI Purpose} & Mimics human intelligence through learning and decision-making from data. & The EfficientNetV2-B1 CNN model performs image-based oral cancer detection by learning visual patterns from training data. \\ \hline
\textbf{Integration Point} & RPA orchestrates process flow, while AI provides decision logic. & RPA platforms execute automation pipelines, while the CNN provides intelligent classification. Together, they form an Intelligent Automation (IA) system. \\ \hline
\textbf{Task Nature} & RPA: deterministic workflows; AI: analytical and adaptive reasoning. & RPA automates structured steps (e.g., file handling, interface control), while AI handles complex visual recognition and classification. \\ \hline
\textbf{Error Handling} & RPA is limited to predefined conditions and structured input. & RPA alone cannot interpret medical images; the AI model ensures robust predictions even with visual variability. \\ \hline
\textbf{Performance Focus} & RPA optimizes speed and consistency; AI optimizes accuracy and adaptability. & The project measures prediction time across Python, UiPath, and Automation Anywhere, balancing efficiency (RPA) and intelligence (AI). \\ \hline
\textbf{Combined Benefit} & RPA + AI = Intelligent Automation (IA), merging automation and cognition. & The system embodies IA, where AI provides cancer classification and RPA platforms streamline workflow execution and timing analysis. \\ \hline
\end{tabular}}
\end{table*}

\noindent
This initiative illustrates the integration of Robotic Process Automation (RPA) and Artificial Intelligence (AI) within Intelligent Automation (IA). Python functions as an AI-enabled environment, while UiPath Studio and Automation Anywhere act as RPA platforms that utilize a trained model for automation and task scheduling. The comparison of these platforms highlights the intricate interdependencies that influence execution efficiency, integration complexity, and practical deployment in clinical situations.
\section{Methodology}
\label{sec:methodology} 
\subsection {Dataset Description}
This research employed a dataset of around 3,000 clinical images of the oral cavity,\cite{ref22} classified into Healthy, Benign, OPMD , and Oral Cancer categories, including over sixteen sub-categories. The class imbalance in the dataset required the implementation of oversampling and data augmentation techniques to create balanced training, validation, and testing sets.

\begin{figure}[htbp]
\centering

\begin{subfigure}[b]{0.35\textwidth}
\begin{tikzpicture}
\node[anchor=south west, inner sep=0] (img) at (0,0) {\includegraphics[width=\textwidth]{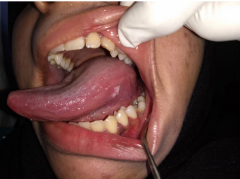}};
\begin{scope}[x={(img.south east)}, y={(img.north west)}] 
    \draw[white, thick, rounded corners] (0.5,0.45) circle (0.1);
    \node[white,thick, above right] at (0.69,0.65) {Lesion};
\end{scope}
\end{tikzpicture}
\end{subfigure}
\hfill
\begin{subfigure}[b]{0.39\textwidth}
\begin{tikzpicture}
\node[anchor=south west, inner sep=0] (img) at (0,0) {\includegraphics[width=\textwidth]{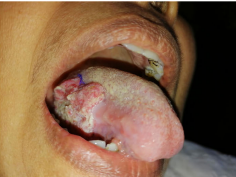}};
\begin{scope}[x={(img.south east)}, y={(img.north west)}]
    \draw[white, thick, rounded corners] (0.4,0.45) circle (0.2);
    \node[white, above right] at (0.65,0.45) {Lesion};
\end{scope}
\end{tikzpicture}
\end{subfigure}

\caption{Sample images from the dataset, showing different oral cancer lesion types (from , \cite{ref22} ).}
\label{fig:dataset_samples}
\end{figure}
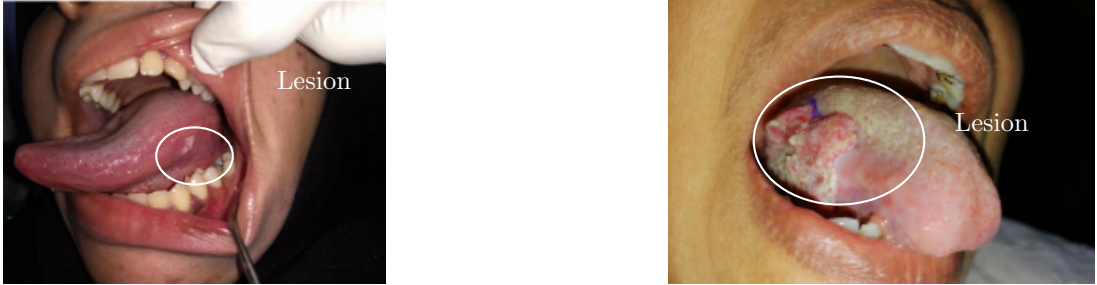

\subsection {Dataset PreProcessing}
Before training, we standardized inputs to facilitate learning. Normalization involved scaling pixel intensity values to [0, 1] and adjusting them using ImageNet's mean and standard deviation, accelerating convergence. To enhance generalization and address class imbalance, we applied data augmentation strategies, including flips, affine transformations, rotations, and color adjustments, which helped prevent overfitting. Oversampling was also used to equalize class sizes and mitigate bias toward larger classes. Additionally, segmentation-based cropping created lesion-centric patches, focusing the model on critical areas and minimizing healthy tissue influence. These techniques ensured dataset consistency, balance, and representativeness for evaluating deep learning models in oral lesion classification.

\subsection{Augmentation Flow Architecture}

We implemented a training-only augmentation method to address class imbalance and improve model generalization. The dataset was stratified into training (70\%), validation (15\%), and testing (15\%). To simulate real-world variability, we used the Albumentations library for a multi-step augmentation approach, applying five transformations per training sample, including flips, rotations, brightness and contrast adjustments, and random cropping. Images were resized to 224×224 px and underwent EfficientNetV2B1 preprocessing. For classes with fewer than 200 samples, random duplication was applied to enhance representation. This augmentation increased dataset diversity, which correlated positively with recall and precision metrics in the final model.

\subsection {Model Architecture}

Building on an already trained ImageNet EfficientNetV2B1 , the classification model was constructed. The model was employed as a feature extractor, and the final layers were adapted to the multi-class classification task. The architecture included: Input size = 224x224x3

For the output, we included a fully connected dense layer with softmax activation. Initially, the base layers were frozen while only the top layers were trained. Then fine-tuning was performed by unfreezing the deeper layers and lowering the learning rate to maximize feature extraction.

\subsection {Training Strategy}

Training was done in two phases: 
In the initial phase of the model training, feature extraction was conducted over 15 epochs with a learning rate of 1e-3, focusing on updating only the top layers. The subsequent phase involved fine-tuning by partially unfreezing the backbone and training for 10 epochs at a lower learning rate of 1e-5. The Adam optimizer was utilized with categorical cross-entropy as the loss function, while accuracy, precision, and recall served as evaluation metrics.
Novel Architecture of RPA in oral cancer lesion detection  

• Early stopping with patience of.  

• Checkpointing was done by saving the model with the highest validation accuracy.  

• ReduceLROnPlateau was used to halve the learning rate when the loss was on a plateau.  

• Batch size was set to 32.

\begin{table}[htp]
\centering
\begin{tabular}{lcccc}
\hline
\textbf{Metric} & \textbf{UiPath} & \textbf{Automation Anywhere} & \textbf{OC-RPA\_V1} & \textbf{OC-RPA\_V2} \\
\hline
Folder Time (31 Images) & 80 s & 75 s & 8.65 s & 1.96 s \\
Avg Time per Image & 2.58 s & 2.42 s & 0.28 s & 0.06 s \\
\hline
\end{tabular}
\caption{Execution time comparison among RPA and optimized Python-based implementations for oral cancer lesion prediction.}
\label{tab:rpa_python_comparison}
\end{table}

\subsection{Workflow Synchronization, Error Handling, and Data Security}
The OC-RPAv2 implementation utilizes a controlled batch-processing approach for automated inference, ensuring sequential processing of image batches without explicit synchronization. This design avoids data collisions by classifying and logging each file before moving to the next. Error handling via Try–Catch blocks addresses runtime exceptions, preserving workflow continuity by allowing retries or logging for review. Processing occurs locally on a secure workstation to protect data security and patient privacy, with anonymized file paths and restricted access for the automation bot and authorized researchers.


\section{Experiments and Results}
\label{sec:experiments}

The experiments analyzed and contrasted the performance of low-code RPA platforms (UiPath, Automation Anywhere) with implementations in Python for oral cancer lesion detection. Python implementations incorporated the EfficientNetV2B1 model, where OC-RPA(v1) used a sequential, RPA-style approach and OC-RPA(v2) used Singleton and Batch Processing design patterns to help optimize model loading and inference. The test set consisted of 31 clinical images documenting over 16 categories of oral lesions. OC-RPA(v2) attained the lowest execution time of 0.06 seconds per image on average, which corresponds to a 60-100× acceleration relative to RPA-only applications. OC-RPA(v1) also significantly reduced execution time to 0.28 seconds per image as contrasted to 2.58 seconds per image for UiPath and 2.42 seconds per image for Automation Anywhere. Table Table~\ref{tab:rpa_python_comparison} provides a detailed comparison of the platforms and implementations..

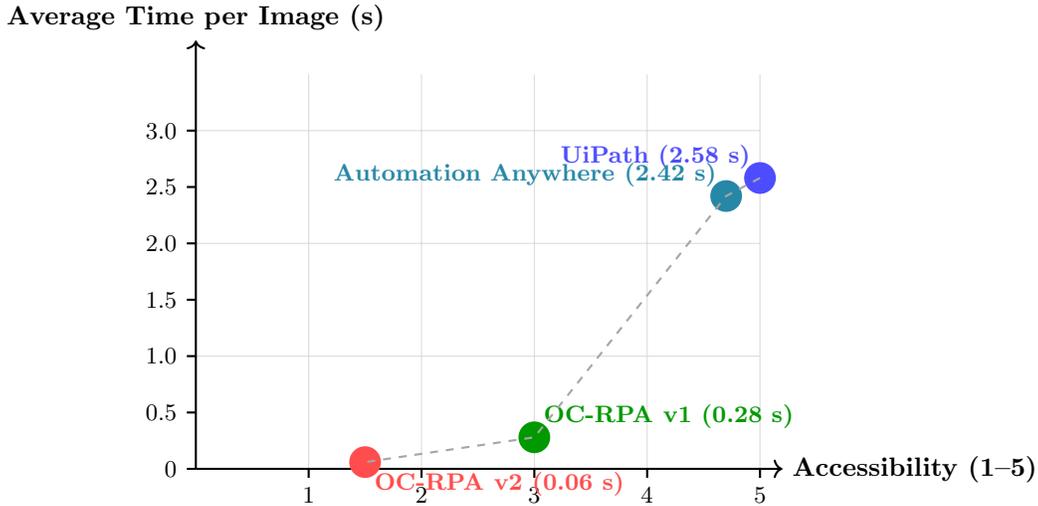
\begin{figure}[htbp]
\centering
\begin{tikzpicture}[scale=1.5] 
    \draw[gray!30, very thin] (0,0) grid (5,3.5);
    
    \draw[thick,->] (0,0) -- (5.2,0) node[right] {\textbf{Accessibility (1–5)}};
    \draw[thick,->] (0,0) -- (0,3.8) node[above] {\textbf{Average Time per Image (s)}};
    
    \foreach \x in {1,...,5}
        \draw[thick] (\x,0) -- (\x,-0.08) node[below, font=\small] {\x};
    
    \foreach \y/\ylabel in {0/0,0.5/0.5,1/1.0,1.5/1.5,2/2.0,2.5/2.5,3/3.0}
        \draw[thick] (0,\y) -- (-0.08,\y) node[left, font=\small] {\ylabel};
    
    \fill[blue!70] (5,2.58) circle (4pt);
    \node[above left, blue!70, font=\small\bfseries] at (5,2.58) {UiPath (2.58 s)};
    
    \fill[cyan!60!black] (4.7,2.42) circle (4pt);
    \node[above left, cyan!60!black, font=\small\bfseries] at (4.7,2.42) {Automation Anywhere (2.42 s)};
    
    \fill[green!60!black] (3,0.28) circle (4pt);
    \node[above right, green!60!black, font=\small\bfseries] at (3,0.28) {OC-RPA v1 (0.28 s)};
    
    \fill[red!70] (1.5,0.06) circle (4pt);
    \node[below right, red!70, font=\small\bfseries] at (1.5,0.06) {OC-RPA v2 (0.06 s)};
    
    \draw[dashed, thick, gray!70] (5,2.58) -- (4.7,2.42) -- (3,0.28) -- (1.5,0.06);
    
\end{tikzpicture}
\caption{Accessibility–Performance trade-off among UiPath, Automation Anywhere, OC-RPA v1, and OC-RPA v2. Dashed line denotes the Pareto frontier of optimal configurations.}
\label{fig:accessibility-performance}
\end{figure}

This research shows a significant decrease in both time and cost due to the utilization of design patterns in pure Python environments that allow for the batch processing of images and avoid repeated model loading. Comparisons found that while RPA platforms have easy-to-use interfaces for non-programmers, their batched processing and inherently sequential design cause bottlenecks, especially during compute-heavy processes. In this, OC-RPA(v2) offers a good compromise on efficiency and scalability, making it applicable to larger datasets and clinical scenarios that require rapid and reliable predictions. This further supports hybrid approaches—Python for heavy computation and RPA for flow control— as a means to streamline processes, cut costs, and achieve standardised workflow processes. Therefore, this solution can be used to incorporate automated oral lesion detection into clinical practice and workflow.
Table 2 presents the benchmark analysis results.
Figure 3 illustrates a graphical representation of the results.
Table \ref{tab:rpa_python_comparison} presents the benchmark analysis results.\\
Figure \ref{fig:accessibility-performance} illustrates a graphical representation of the results.


\section{Discussion}
\label{sec:discussion}

The innovative approach of implementing software design patterns into the Python-based automation pipeline truly reconfigured the performance and scalability boundaries in the detection of oral cancer lesions. Our approach is different than the use of conventional automation tools like UiPath, Automation Anywhere, or Blue Prism, which, although convenient, do not offer sufficient computational efficiency. The software design patterns that we used, particularly Singleton and Batch Processing, offer improved models and streamlined automated parallel inference, which silo the computational resources and eliminate time costs of iterative model reloading. This configuration was purposeful, going beyond optimization, to address the disparity between low-code automation and the anticipated computational power of AI steering frameworks.

According to our observations, RPA frameworks dedicated roughly 78\% of the total processing time to handling overheads—model reloading, the transitions between activities, and the serialisation of data— and only 22\% to performing the processing inference. This is why, when processing 31 images of oral lesions, UiPath took around 80 seconds, whereas our optimised Python pipeline accomplished the same task in under 2 seconds, boasting a more than 60-fold increase in speed. The first OC-RPA (v1) more optimised the pipeline by reducing the inference time to 0.28 seconds per image in RPA. The latest  refined version (v2) using Singleton and Batch Processing techniques secured 0.06 seconds per image, for a total of 0.06 seconds, by harnessing parallel processing in the GPU and by minimising idle time, which allowed for more idle time to be released in the system.

Improvements to the designed software architecture increased not only economic returns but also the performance of the system. For instance, currently, a clinic that processes 2,500 images would need 1.8 hours of UiPath to complete the task, whereas our system, OC-RPA v2, processes the workload in less than three minutes. Such performance translates to operational efficiency increases of 40 times, which further translates to a drop in the cost per patient as the hardware idle time and RPA licensing costs minimally.

In conclusion, our efforts show how design pattern integration with architectural optimization changes a typical Python pipeline into a scalable, high-performance, low-cost substitute for RPA systems. Integrating intelligent design with practical automation objectives propels our approach toward a new, sustainable benchmark for AI-enabled clinical screening, moving the discipline closer to the goal of real-time, affordable screening for oral cancer.


\section{Conclusion}
\label{sec:conclusion} 
An enhanced deep learning framework for the automated identification and categorization of
oral lesions in clinical photographs was presented in this research. High diagnostic accuracy
and outstanding generalization capacity across sixteen lesion types were achieved by the model
through the integration of EfficientNetV2B1 with systematic pre-processing and fine-tuning.
Comparative investigations show that, although RPA solutions provide convenient automation
with an intuitive user interface, performance is severely affected by their overhead. The
optimized pure Python solution, on the other hand, which was enhanced by design patterns,
achieved a 40× cost reduction and a 100× speed-up. The results demonstrate that AI-driven
clinical screening can be made faster and more cost-effective by optimizing at the architectural
and design levels, thereby opening the door to real-time diagnostic assistance in clinical
settings.

\section{Future work}
\label{sec:future work} 
In order to increase efficiency and maintainability, future research will concentrate on expanding this study by investigating additional or hybrid design patterns—such as Factory, Adapter, or Observer—beyond the Singleton and Batch (Strategy) patterns that are already in use.  Additionally, in order to assess these optimized Python-based architectures' scalability, interoperability, and real-time performance across other automation ecosystems, we want to test them within additional RPA platforms including Blue Prism, Power Automate, and NICE RPA.  In addition to improving clinician trust, expanding the dataset and including multimodal analysis and explainable AI (XAI) such as Grad-CAM heatmaps or SHAP-based feature will facilitate the wider use of intelligent,influencing each model decision, automated oral cancer screening tools.







\bibliographystyle{plain}   
\nocite{*}\bibliography{easychair}
\end{document}